\documentclass{article}

\usepackage{arxiv}

\usepackage[utf8]{inputenc} 
\usepackage[T1]{fontenc}    
\usepackage{hyperref}       
\usepackage{url}            
\usepackage{booktabs}       
\usepackage{amsfonts}       
\usepackage{nicefrac}       
\usepackage{microtype}      
\usepackage{lipsum}		
\usepackage{graphicx}
\usepackage{natbib}
\usepackage{doi}
\usepackage{mathtools}

\usepackage{multirow}
\usepackage[table,xcdraw]{xcolor}
\usepackage{hyperref}

\newcommand{\wrt}{{\it w.r.t. }}    
\newcommand{\eg}{\emph{e.g.}, }     
\newcommand{\ie}{\emph{i.e.}, }     
\newcommand{\etal}{\emph{et al.} }   

\newcommand{\focus}{\textit{Focus} }

\title{Focus! Rating XAI Methods and Finding Biases}

\author{ \href{https://orcid.org/0000-0002-8819-6735}{\includegraphics[scale=0.06]{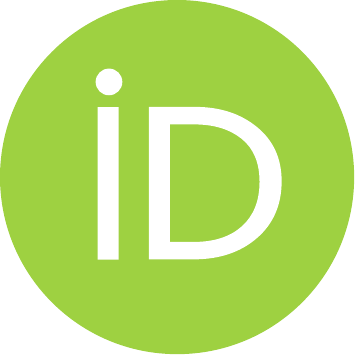}\hspace{1mm}Anna Arias-Duart} \\
	Barcelona Supercomputing Center (BSC) \\
	\texttt{anna.ariasduart@bsc.es} \\
	\And
	\href{ https://orcid.org/0000-0002-8329-0343}{\includegraphics[scale=0.06]{orcid.pdf}\hspace{1mm}Ferran Parés} \\
	Barcelona Supercomputing Center (BSC)\\
	\texttt{ferran.pares@bsc.es} \\
	\And
	\href{ https://orcid.org/0000-0001-6732-5641}{\includegraphics[scale=0.06]{orcid.pdf}\hspace{1mm}Dario Garcia-Gasulla} \\
    Barcelona Supercomputing Center (BSC) \\
	\texttt{dario.garcia@bsc.es} \\
	\And
	\href{https://orcid.org/0000-0003-4514-6145}{\includegraphics[scale=0.06]{orcid.pdf}\hspace{1mm}Víctor Giménez Ábalos} \\
    Barcelona Supercomputing Center (BSC) \\
	\texttt{victor.gimenez@bsc.es} \\
}

\date{}

\renewcommand{\shorttitle}{Focus! Rating XAI Methods and Finding Biases}

\hypersetup{
pdftitle={Focus! Rating XAI Methods and Finding Biases},
pdfsubject={q-bio.NC, q-bio.QM},
pdfauthor={Anna Arias-Duart, Ferran Pares, Dario Garcia-Gasulla},
pdfkeywords={First keyword, Second keyword, More},
}

\begin{document}
\maketitle

\begin{abstract}
AI explainability improves the transparency of models, making them more trustworthy. Such goals are motivated by the emergence of deep learning models, which are obscure by nature; even in the domain of images, where deep learning has succeeded the most, explainability is still poorly assessed.
In the field of image recognition many feature attribution methods have been proposed with the purpose of explaining a model’s behavior using visual cues. However, no metrics have been established so far to assess and select these methods objectively. In this paper we propose a consistent evaluation score for feature attribution methods---the \textit{Focus}---designed to quantify their coherency to the task. While most previous work adds out-of-distribution noise to samples, we introduce a methodology to add noise from \textit{within} the distribution. This is done through mosaics of instances from different classes, and the explanations these generate. On those, we compute a visual pseudo-precision metric, \textit{Focus}. First, we show the robustness of the approach through a set of randomization experiments. Then we use \focus to compare six popular explainability techniques across several CNN architectures and classification datasets. Our results find some methods to be consistently reliable (LRP, GradCAM), while others produce class-agnostic explanations (SmoothGrad, IG). Finally we introduce another application of \textit{Focus}, using it for the identification and characterization of biases found in models. This empowers bias-management tools, in another small step towards trustworthy AI.
\end{abstract}

\section{Introduction}\label{sec:introduction}

Explainability has become a major topic of research in Artificial Intelligence (AI), aimed at increasing trust in models such as Deep Learning (DL) networks. However, trustworthy models cannot be achieved with explainable AI (XAI) methods unless the XAI methods themselves can be trusted. This necessity gave rise to the assessment of XAI methods.

To evaluate XAI methods one may assess interpretability, a \textit{qualitative} measure of how understandable an explanation is to humans \cite{gilpin2018explaining}. While this is important to guarantee the proper interaction between humans and the model, interpretability generally involves end-users in the process \cite{mohseni2018multidisciplinary}, inducing strong biases. In fact, a qualitative evaluation alone cannot guarantee coherency to reality (\textit{i.e.,} model behavior), as false explanations can be more interpretable than accurate ones. To enable trust on XAI methods, we also need \textit{quantitative} and objective evaluation metrics, which validate the relation between the explanations produced by the XAI method and the behavior of the trained model under assessment.

\begin{figure}[t]
\centering
\setlength\tabcolsep{1pt}
\begin{tabular}{cccc}
\includegraphics[width=0.19\textwidth]{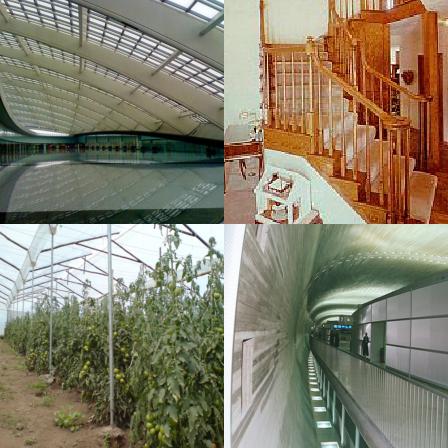} &
\includegraphics[width=0.19\textwidth]{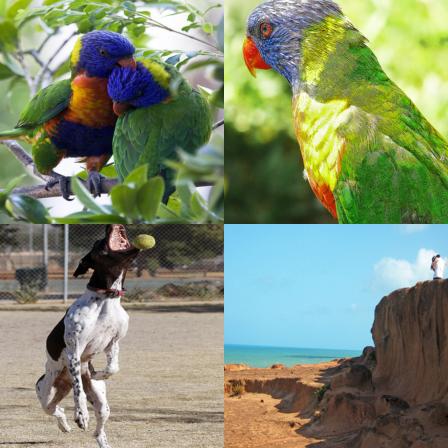} &
\includegraphics[width=0.19\textwidth]{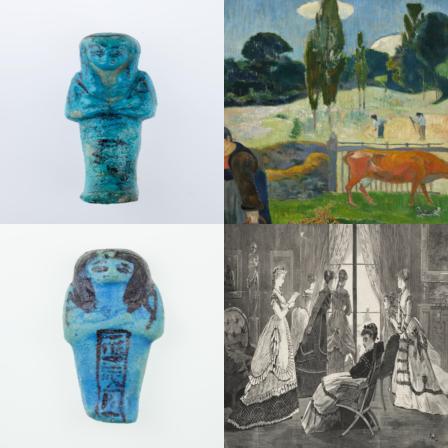}\\
(a) MIT67 & (b) ILSVRC2012 & (c) MAMe  \\
\\
\includegraphics[width=0.19\textwidth]{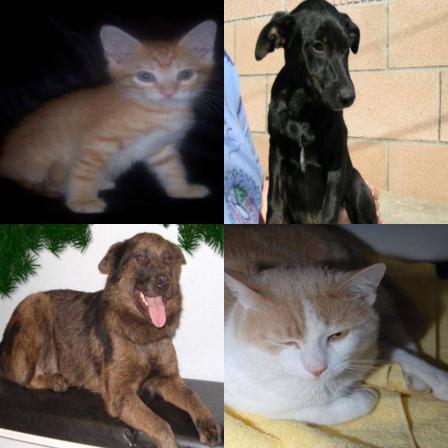} &
\includegraphics[width=0.19\textwidth]{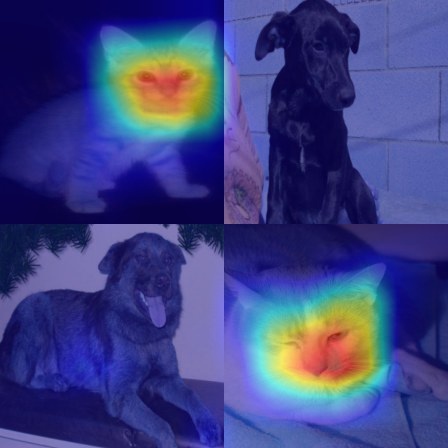} &
\includegraphics[width=0.19\textwidth]{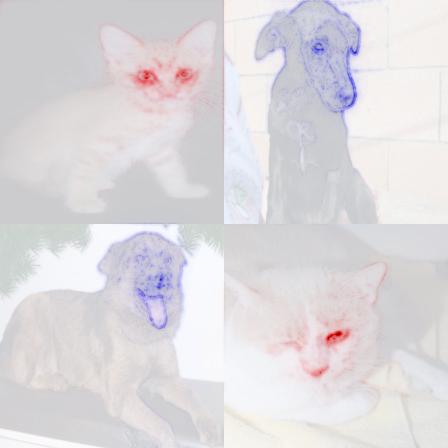}\\
(d) Dogs vs. Cats & (e) GradCAM & (f) LRP \\
\\
\end{tabular}
\caption{First row: sample of mosaics used by the evaluation methodology, obtained for: (a) MIT67 (b) ImageNet (ILSVRC2012) (c) MAMe. Second row: example of input mosaic from Dogs vs. Cats (d), and the explanations obtained by GradCAM (e) and LRP (f) for the target class \textit{cat}.}\label{fig:mosaics}
\end{figure}

The challenge of quantitatively evaluating XAI methods lies in the absence of a ground truth: we cannot be sure of what a DL method is doing unless we understand the model parametrization itself (at which point we would not need a XAI method). Nonetheless, we still want to approximate the faithfulness \cite{Selvaraju_2019} of XAI methods \wrt the underlying model, as this allows us to discern between accurate and misleading explanations. In this paper we propose a novel metric for that purpose, demonstrating its use on the evaluation of \textit{feature attribution} methods when applied on image classification models. 

The evaluation of XAI faithfulness is typically done by quantifying the change produced in the explanation when noise is added to the explained sample. This is necessary because no assumptions can be made regarding the \textit{faithfulness} of samples without noise: explanations apparently inappropriate (\eg the background of a central object instead of object itself) may be an accurate portrait of the model's behavior, following a bias found and learnt from the data. The most popular approach to add noise is to visually alter samples \cite{bach2015pixel, chattopadhay2018grad}. However, disturbed images become images outside of the original data distribution, which reduces the reliability of the analysis because of the effect it may cause on the activations of the model (\ie are bad explanations caused by a bad method or by the corruption inserted into the samples?). 

In this paper we propose a novel evaluation score for \textit{feature attribution} methods, described in \S\ref{sec:methodology}. Our input alteration approach induces in-distribution noise into samples, that is, alterations on the input which correspond to visual patterns found within the original data distribution. To do so we modify the context of the sample instead of the content, leaving the original pixels values untouched. In practice, we create a new sample, composed of samples of different classes, which we call a \textit{mosaic image} (see examples in Figure~\ref{fig:mosaics}). Using \textit{mosaics} as input has a major benefit: each input quadrant is an image from the original distribution, producing blobs of activations in each quadrant which are consequently coherent. Only the pixels forming the borders between images, and the few corresponding activations, may be considered out of distribution.

By inducing in-distribution noise, \textit{mosaic images} introduce a problem in which XAI methods may objectively err (\ie focus on something it should not be focusing on). On those composed mosaics we ask a XAI method to provide explanation for just one of the contained classes, and follow its response. In a sort of eye-tracking game, we measure how much of the explanation generated by the XAI is located on the areas corresponding to the target class, quantifying it through the \focus score. This score allows us to compare methods in terms of explanation precision, evaluating the capability of XAI methods to provide explanations related to the requested class (see \S\ref{sec:results_and_discussion} for the comparison of six XAI methods).

Using \textit{mosaics} has another benefit. Since the noise introduced is in-distribution, the explanation errors identify and exemplify biases of the model. This facilitates the elimination of biases in models and datasets, potentially resulting in more reliable solutions. We illustrate how to do so in \S\ref{sec:bias_detection}.

\section{Related Work} \label{sec:related_work}

The evaluation of XAI faithfulness in current literature can be divided into two broad groups: qualitative and quantitative methods. On one side, qualitative evaluations are based on assumptions induced by the human understanding of perception~\cite{Selvaraju_2019, ribeiro2016should}, hence an evaluation built on top of human cognitive biases, may not be necessarily aligned with the learning paradigm of DL. In contrast, quantitative evaluations avoid such human biases by excluding the human from the XAI assessment process.

Quantitative evaluations have a primary barrier to overcome: the lack of a ground truth specifying what defines a correct explanation. Instead, these methods introduce noise to evaluate the output, assuming certain properties on their expected response. Such noise can be introduced both on data and model parametrization.

We can separate noise inducing evaluation methods based on their generated response. While some produce categorical evaluation, others generate numerical ones. Works proposing categorical evaluations define axioms or tests that XAI methods must satisfy or fulfill. One of these works~\cite{sundararajan2017axiomatic} discusses three axioms for XAI methods: \textit{Sensitivity}, \textit{Implementation invariance} and \textit{Completeness}. \textit{Sensitivity} checks that irrelevant features have no explanation attributed, \textit{Implementation invariance} checks that functionally similar models produce equivalent attributions, and \textit{Completeness} is satisfied when the difference between the sum of the attributions of an input and a baseline is equal to the change of the output. In \cite{adebayo2020sanity}, two types of tests are proposed: the \textit{model parameter randomization} test and the \textit{data randomization} test. The first aims to prove that if an explanation depends on the model, a randomized model should produce a different explanation. The second test checks if there exists a relation between the explanation and the labels, that is, if a regular dataset produces different results than one with randomly permuted labels. 

While categorical evaluations serve for validating whether the XAI methods fulfill or not certain properties, they are limited in terms of comparison and ranking purposes. Numeric evaluations are more informative to that end, providing comparable scores that can be easily ranked. Examples of numeric evaluations are the Pixel Flipping algorithm \cite{bach2015pixel} and the Average Drop \% metric \cite{chattopadhay2018grad}. The first performs a semi-quantitative analysis by perturbing pixels from patches with the greatest relevance, and then assessing the impact on the prediction score. The second measures the drop percentage of the score when only the part of the image with attribution is shown \wrt the score with the full image. These numeric evaluations rely on disturbing input images. As said before, disturbed images fall out of the original distribution, reducing the reliability of the following analysis due to its effect on produced model activations. 

A few XAI assessment methods use images from the original distribution without any perturbation on them. Since these methods lack a source of noise, they require of an assumption to numerically evaluate the produced explanations. Examples of these methods are those working with manually generated ground truth regions, assuming that the relevance produced by XAI methods should fall inside regions corresponding to a target class. In the work of Zhang \etal \cite{zhang2018top}, authors introduce the \textit{pointing game} technique to evaluate if the point of maximum relevance lies on the object of the target class. Similarly, Selvaraju \etal evaluate the localization capacity of methods by drawing bounding boxes from the explanation heatmap and calculating the error \wrt the correct bounding box \cite{Selvaraju_2019}. However, as pointed out by different works \cite{sundararajan2017axiomatic, Samek_2019}, the premise that objects, or any other part of the input, are the only relevant feature for the prediction cannot be presumed.

In this context, we define the \focus score. A quantitative, numerical, in-distribution noise inducing method to assess XAI methods and AI models (see \S\ref{sec:results_and_discussion} and \S\ref{sec:bias_detection}). It is based on compositions of images from the dataset as a source of noise. Since each quadrant of the mosaic contains an undisturbed image, the network activations in each quadrant will fall inside the original distribution. Additionally, the \focus score does not expect relevance to be only centered on a pre-defined region, avoiding assumptions regarding the localization of explanation within each quadrant.

\section{Methodology} \label{sec:methodology}
In this section we define a novel metric---the \textit{Focus}--- intended to assess the explanations produced by \textit{feature attribution} methods. This score involves three elements: an explainability method (\S\ref{sec:xai_methods}), a trained model (\S\ref{sec:training_set_up}), and a set of mosaic samples (\S\ref{sec:mosaic_construction}). In the following subsections we discuss these in detail, before defining the \focus metric itself (\S\ref{sec:focus_metric}).

\subsection{Explainability methods}\label{sec:xai_methods}

Throughout the paper we use and evaluate six \textit{feature attribution} methods:
\begin{itemize}
    \item Gradient-weighted Class Activation Mapping (GradCAM)~\cite{Selvaraju_2019}, based on the implementation of Gildenblat \etal \footnote{\url{https://github.com/jacobgil/pytorch-grad-cam}\label{fn:gradcamgithub}}. We compute the gradients of the logits of the class \wrt the feature maps of the final convolutional layer. That is, the 5th layer for AlexNet, the 13th for VGG16 and the last layer from the 5th block for ResNet-18 (also known as block E).
    \item Layer-wise Relevance Propagation (LRP)~\cite{bach2015pixel}, based on the implementation of Nam \etal \cite{nam2019relative}. On the first layer we use the $z^B$-rule \cite{montavon2017explaining}, on fully connected layers the LRP-$\epsilon$ \cite{bach2015pixel}, and on convolutional layers the LRP-$\alpha \beta$ \cite{bach2015pixel} with $\alpha=1$ and $\beta=0$.
    \item SmoothGrad~\cite{smilkov2017smoothgrad}, based on the implementation of Nakashima \etal \footnote{\url{https://github.com/kazuto1011/grad-cam-pytorch}}. Explanations are obtained computing the gradient of the specific class score \wrt the input pixels and adding small perturbations on the input image (in our case Gaussian Noise). 
    \item LIME \cite{ribeiro2016should}, based on the implementation of Tulio \etal \footnote{\url{https://github.com/marcotcr/lime}}.
    Each explanation is computed considering 1,000 samples and the final explanation only includes the five top features, that is, the five most relevant superpixels. 
    \item GradCAM++ \cite{chattopadhay2018grad}, based on the implementation of Gildenblat \etal \textsuperscript{\ref{fn:gradcamgithub}}. We use the last convolutional layer to compute the GradCAM++ explanations.
    \item Integrated Gradients (IG)\cite{sundararajan2017axiomatic}, based on the 
    implementation of Kokhlikyan \etal \cite{kokhlikyan2020captum}. We use the black image as the baseline image and 30 steps to approximate the integral.
\end{itemize}

For all \textit{feature attribution} methods we skip their custom post-processing for visualization purposes.

\subsection{Models}\label{sec:training_set_up}

To run a XAI method one needs a model to explain. One generated from an architecture trained on a dataset, through a specific training configuration. In our experiments, we use the following:

\begin{itemize}
    \item \textbf{Architectures: }AlexNet \cite{krizhevsky2012imagenet}, VGG16 \cite{simonyan2014very} and ResNet-18 \cite{he2016deep}.
    \item \textbf{Datasets: }the Dogs vs. Cats\footnote{\url{https://www.kaggle.com/c/dogs-vs-cats/overview}}, the Museum Artworks Medium dataset (MAMe) \cite{pares2021mame}, the MIT67 \cite{quattoni2009recognizing} and the ILSVRC 2012 \cite{russakovsky2015imagenet} (hereafter ImageNet).
\end{itemize}

\textbf{Training configurations:} During training, AMSGrad~\cite{reddi2019convergence} is used for optimizing weights and data augmentation is performed. Code needed to replicate trainings and experiments of this paper can be found in~\footnote{\url{https://github.com/HPAI-BSC/Focus-Metric}
}. For the ImageNet dataset, we use the pre-trained models in the subpackage \textit{torchvision.models}\footnote{\label{alexnet}\url{https://download.pytorch.org/models/alexnet-owt-4df8aa71.pth}}\textsuperscript{,}\footnote{\label{vgg16}\url{https://download.pytorch.org/models/vgg16-397923af.pth}}\textsuperscript{,}\footnote{\label{resnet18}\url{https://download.pytorch.org/models/resnet18-5c106cde.pth}}. For Dogs vs. Cats and MAMe datasets, we take the ImageNet pre-trained models and perform training on top of them. Finally, in the case of the MIT67 dataset, we train the model on top of pre-trained Places365-Standard dataset \cite{zhou2017places} (models available in the official repository \footnote{\label{places}\url{https://github.com/CSAILVision/places365}}).

\subsection{Mosaic construction}\label{sec:mosaic_construction}

The last element required to compute the \focus metric is the mosaic, an image composed by four different samples disposed in a two by two grid. Samples from the training set are never used for mosaics. 

To formalize mosaics, and later \textit{Focus}, let us define a dataset $\mathbb{D}$ composed by a set of images $\mathbb{I}=\{img_1, img_2, ... , img_N\}$ and a set of classes $\mathbb{C}=\{c_1, c_2, ... , c_K\}$, where $N$ is the number of total images and $K$ is the number of total classes. Every image in $\mathbb{I}$ has assigned a unique class from $\mathbb{C}$: $c(img)$. From here we build a set of mosaics $\mathbb{M} = \{m_1, m_2, ... , m_J\}$ where $J$ is the total number of mosaics in $\mathbb{M}$. A mosaic $m$ is composed by four images $m = \{img_1, img_2, img_3, img_4\}$ and characterized by a target class $tc = c(m)$, the specific class the XAI method is expected to explain. While two images of the mosaic belong to the target class $c(img_1) = c(img_2) = c(m)$, the other two are randomly chosen among the rest of classes $c(img_3) \neq c(m); c(img_4) \neq c(m)$. Mosaics are implemented as two by two, non-overlapping grid, with the position of each image being random. Samples of mosaics from different datasets can be seen in Figure~\ref{fig:mosaics}.

For the sake of keeping the same resolution of the visual patterns seen by models during their training, and thus keeping most of the noise added within the training distribution, all XAI evaluation experiments use 448$\times$448 mosaics. That is, four times the size of the inputs the models were trained with. Noticeably, both AlexNet and VGG16 architectures were not input-agnostic when originally proposed, being limited by design to an input size of 224$\times$224 pixels. Nowadays, these architectures employ an Adaptive Pooling Layer to circumvent this problem. 

\subsection{The Focus metric}\label{sec:focus_metric}

Before starting with the \textit{Focus}, let us introduce its foundations as well as its motivation. When a \textit{feature attribution} method is applied to an image to explain the model's prediction regarding a chosen class, it typically produces a map from pixels to real values, referred to as relevance. While some \textit{feature attribution} methods also provide negative relevance, this is not generalized. For the scope of this paper we \textit{focus} on positive relevance only. For XAI methods providing both positive and negative relevance, only the positive relevance is used, while negative values are treated as 0.

Intuitively, the output of a method is reliable (but not necessarily understandable) when higher values of relevance lie on pixels of the image that are visual evidence toward the chosen class. We consider \textit{visual evidence} any set of pixels used by the model to distinguish the chosen class from any other class of the task. To formalize this we introduce a probability distribution $\mathcal{P}_{tc}$ over all possible pixels given a target class $tc$. The probability of sampling a pixel from $\mathcal{P}_{tc}$ is proportional to the pixel's relevance toward $tc$ attributed by an explainability method $\mathcal{A}$ and a model $\theta$. Then, we define the formal reliability $Re(\mathcal{A}, \theta, tc)$ as the probability that a pixel sampled from the distribution $\mathcal{P}_{tc}$ lies within visual evidence corresponding to $tc$.

The definition of $Re(\mathcal{A}, \theta, tc)$ over a method-model-class triplet can be extended to evaluate a method-model pair as $Re(\mathcal{A}, \theta)$. To do so, we take the expectancy of reliability over all classes $\mathbb{C}$: $Re(\mathcal{A}, \theta) = \mathop{\mathbb{E}}_{tc \in \mathbb{C}}[Re(\mathcal{A}, \theta, tc)]$. More accurate models and better \textit{feature attribution} methods will result in $Re(\mathcal{A}, \theta)$ values closer to 1. The lower bound of $Re(\mathcal{A}, \theta)$ is the probability that any pixel lies within evidence, which is proportional to the number of pixels lying on visual evidence.

In order to obtain the $Re(\mathcal{A}, \theta)$ metric, we would require a ground truth of which pixels are evidence toward a class. A way to bypass this limitation is to take the assumption that evidence toward a class is more prevalent in images labelled with that class, this being the main assumption of the proposed approach. We thus define the \focus as an estimator of the reliability computed over a dataset. The \focus evaluates the expected probability that a pixel sampled from $\mathcal{P}_{tc}$ lies on an image of $tc$. Notice the \focus underestimates the reliability, as evidence toward a class can be present on samples of a different class of the dataset. We leverage this to our advantage in \S\ref{sec:bias_detection}, using it to detect biases in models and dataset (be it desirable or undesirable biases).

Since this new score only requires image labelling instead of pixel labelling, we transform the dataset into a set of mosaics as introduced in \S\ref{sec:mosaic_construction}. As such, we compute \focus on subsets of four images (\ie each image composing the mosaic is labeled) to estimate the \focus of a method and a model on the whole dataset.

In this context, the \focus metric estimates the reliability of XAI method's output as the probability of the sampled pixels lying on an image of the target class of the mosaic $c(m)$. This is equivalent to the proportion of positive relevance lying on those images:
\begin{equation}
    F_{\mathcal{A}, \theta}(m) = \frac{R_{c(m)}(img_1) + R_{c(m)}(img_2)}{R_{c(m)}(m)}
\end{equation}
where $R_{c}(r)$ is the sum of positive relevance toward class $c$ on the region of the mosaic $r$.

This probability can be interpreted as a precision of the relevance. In an sort of eye-tracking game, the \focus metric asks to the XAI method ``\textit{Why does mosaic $m$ belong to class $c(m)$?}'' on a mosaic $m$ which contains both samples belonging and not belonging to the target class $c(m)$. Given the previous question and a good underlying model, a reliable \textit{feature attribution} method should be able to concentrate most of its explanation relevance on the two appropriated images of the mosaic (\ie $img_1$ and $img_2$).

As explainability becomes more reliable, the \focus will grow. As with reliability, the theoretical upper-bound of the \focus score is $1$, but this is unrealistic: \textit{visual evidence} of a class appearing exclusively on images of that class is seldom true. On the other hand, in the case of uninformed relevance attribution (\ie unreliable explanations), the expected value of \focus is $0.5$, since the probability of picking a pixel of the correct class is just the prior probability of picking one of the pixels of $img_1$ or $img_2$, which amount to half of the total pixels in the mosaic.

\section{Randomization test}

Current evaluations of XAI methods frequently rely on qualitative assessments. These include humans in the loop, thus introducing a significant subjective bias. This is further complicated by the fact that XAI methods are typically designed to focus on prominent, central and/or high contrast areas on the input. When this happens, a XAI method may become more dependant on the input sample than on the underlying model supposedly generating the explanations. To verify this is not an issue for the \focus score, we run a set of randomization tests. 

\begin{figure}[t]
    \centering
    \includegraphics[width=0.46\textwidth]{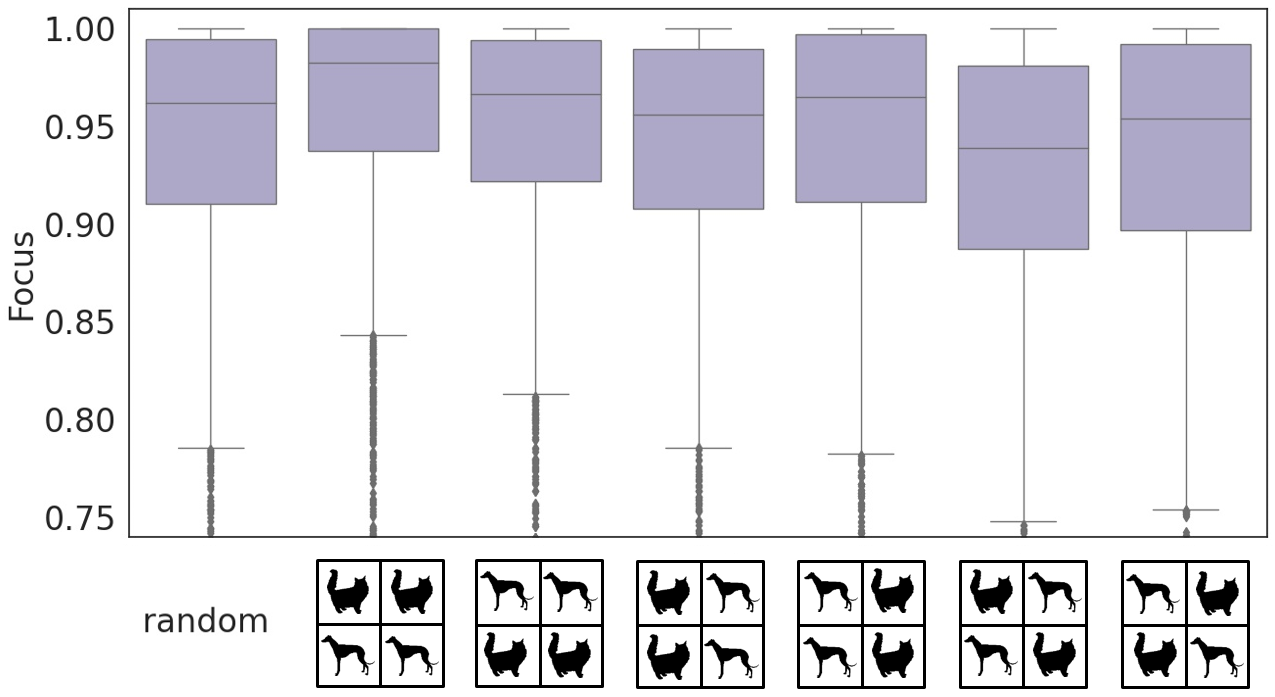}
    \caption{\focus obtained by GradCAM on a VGG16 trained for Dogs vs. Cats dataset (pre-trained on ImageNet), using different mosaic configurations. Each box plot shows the distribution of \focus obtained from evaluating 2,812 samples
    for each configuration (the cat being the target class).
    }
    \label{fig:mosaic_configuration}
\end{figure}

First, we conduct a randomization experiment to assess and decide the exact position of the target class ($tc$) images within the two by two grid of the mosaic. This experiment uses GradCAM on top of a VGG16 model trained for the Dogs vs. Cats dataset (pre-trained on ImageNet). The six possible configurations of the two by two grid were tested, plus a seventh for random positioning. For each configuration 2,812 mosaics were created, using \textit{cat} class as $tc$. The resulting \focus distributions are shown in Figure \ref{fig:mosaic_configuration}. Clearly, the positioning of target samples has an effect on the \focus distribution. Configurations where the two target class images ($img_1$ and $img_2$) are arranged contiguously tend to be better. While this may be partially the result of explanation relevance spilling over samples, it happens more prominently when correct samples are placed on top. Meanwhile, the left-right configurations show a smaller gain when placing the correct samples on the right. Since we cannot guarantee that these properties will hold among target classes, datasets or models, we decide to use a sampling approach hereafter. That is, the exact position samples within the composed grid is chosen randomly for every mosaic.

\begin{figure}[t]
    \centering
    \includegraphics[width=0.46\textwidth]{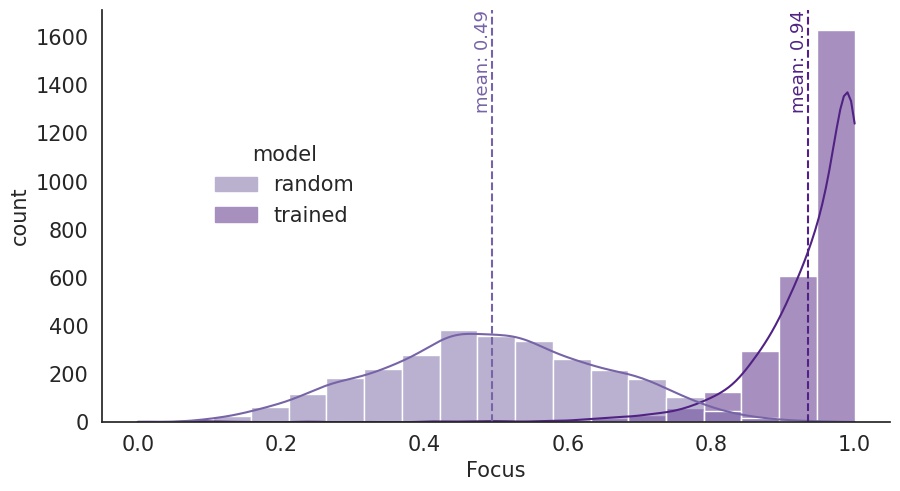}
    \caption{Histogram of \focus scores obtained by GradCAM from 2,812 mosaics, using a VGG16 trained on Dogs vs. Cats and a randomized VGG16 model. The corresponding PDF estimation is represented by a contour line on top.}
    \label{fig:trained_random}
\end{figure}

The second randomization test aims at evaluating the effect of model randomization on the \focus score. For that, we start using two different models. A VGG16 pre-trained on ImageNet and then trained for Dogs vs. Cats, and a totally randomized VGG16 model. The experiment computes the \focus metric on the cat target class ($tc=cat$) for the 2,812 mosaics with random layout. The distribution of \focus achieved by GradCAM on both models is shown as histograms in Figure \ref{fig:trained_random}. While the mean of the \focus obtained with the pre-trained model reach a remarkable 0.94, the random model mean score is 0.49, that is roughly 50$\%$ of the relevance lays on the wrong class quadrants. To take the randomization analysis further, we replicate the experiment of Adebayo \etal\cite{adebayo2020sanity}. In it, the authors qualitatively pointed at how visual explanations can be compelling to the eye even when randomizing one or more layers of the underlying model. In this experiment, layers are randomized in cascade, starting with only the top layer, and increasingly randomizing more layers one by one until obtaining a fully randomized model. We use GradCAM on InceptionV3 \cite{szegedy2016rethinking} (like \cite{adebayo2020sanity}) adding as well VGG16 and ResNet-18. Our results are straight-forward: simply randomizing the top layer (or any other set of layers) makes the \focus drop to a 50$\%$ mean, the same score obtained by a purely random XAI method. This illustrates how resistant the \focus score is to misleading explanations.


\section{Evaluation of XAI methods}
\label{sec:results_and_discussion}

\begin{figure*}[!htp]
\centering
\setlength\tabcolsep{2pt}
\begin{tabular}{cc}
\includegraphics[width=0.49\textwidth]{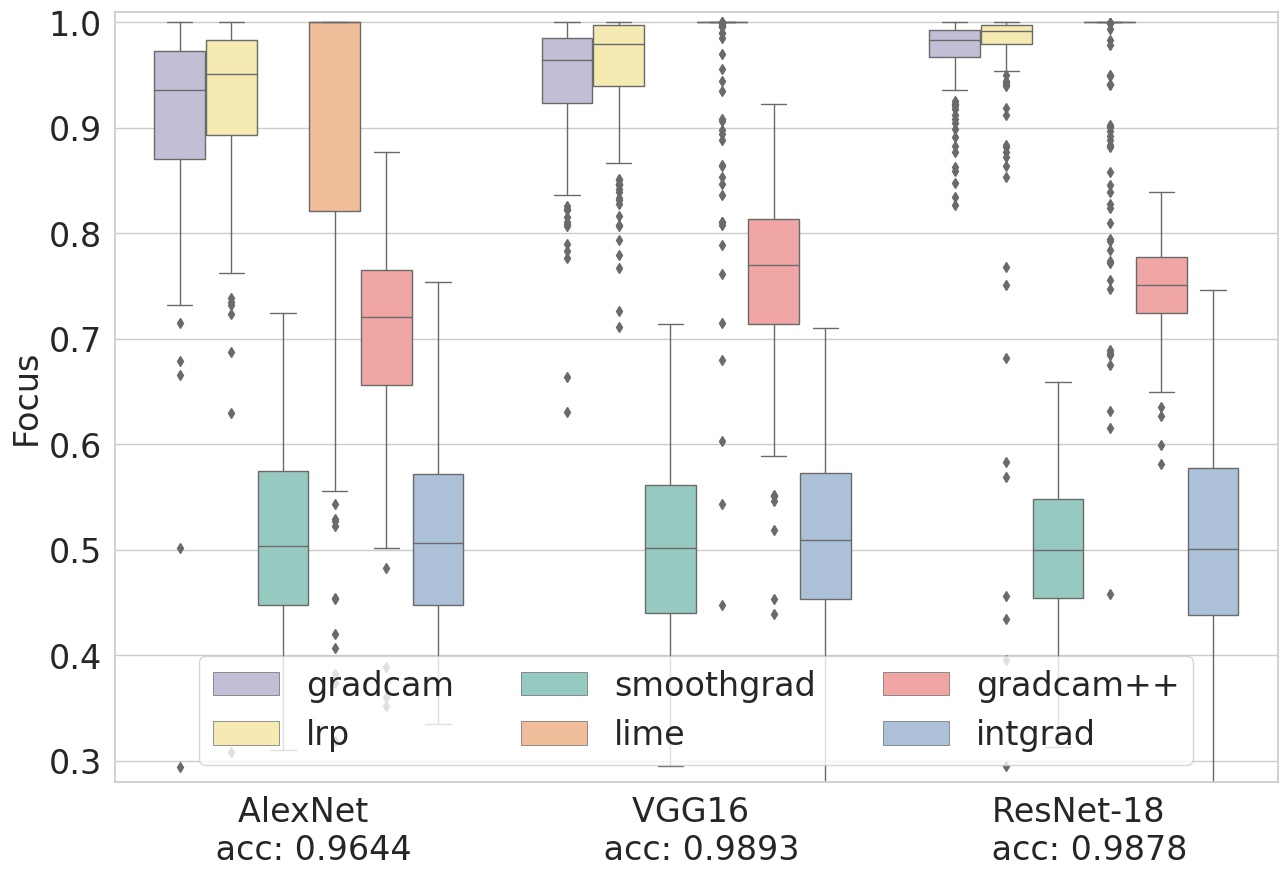} &
\includegraphics[width=0.49\textwidth]{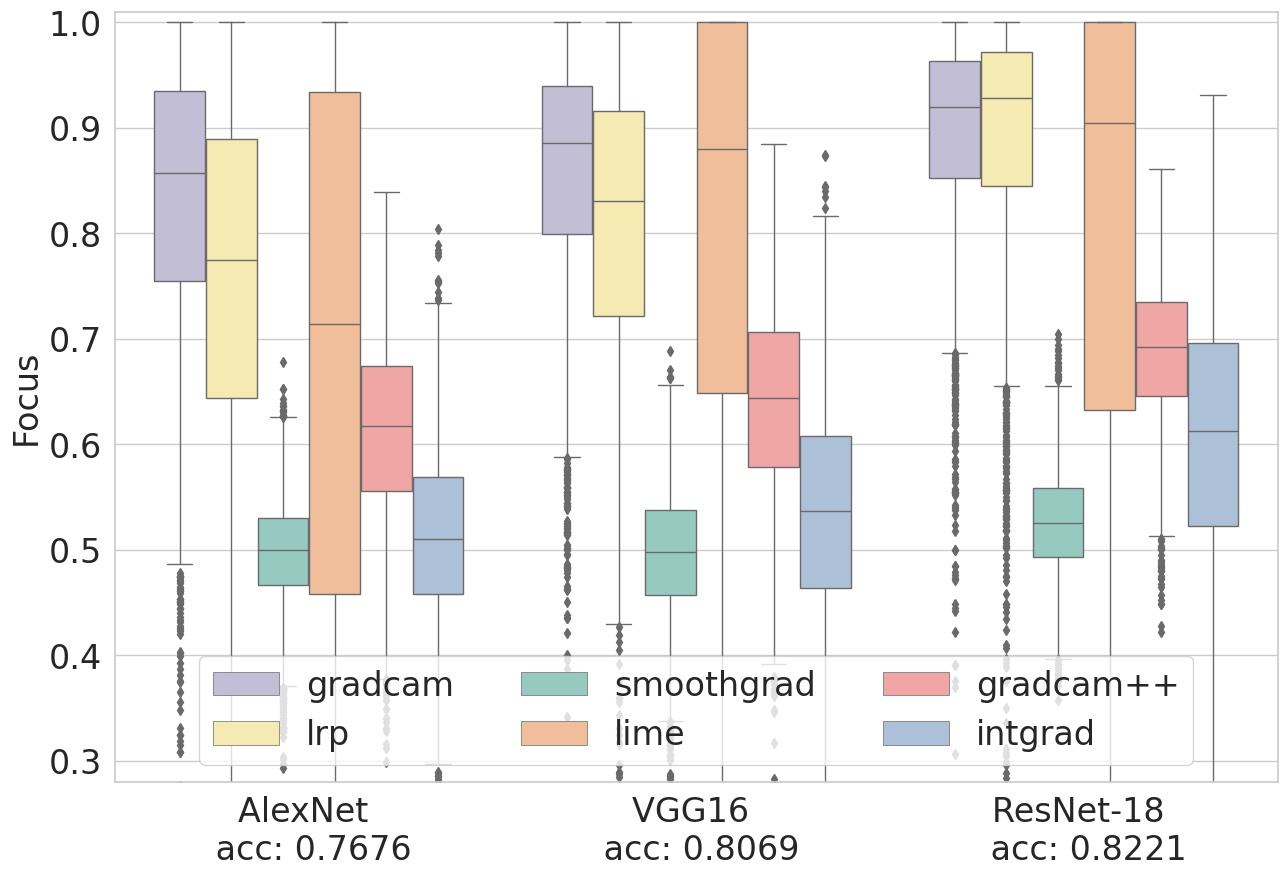} \\
(a)  & (b) \\
\\
\includegraphics[width=0.49\textwidth]{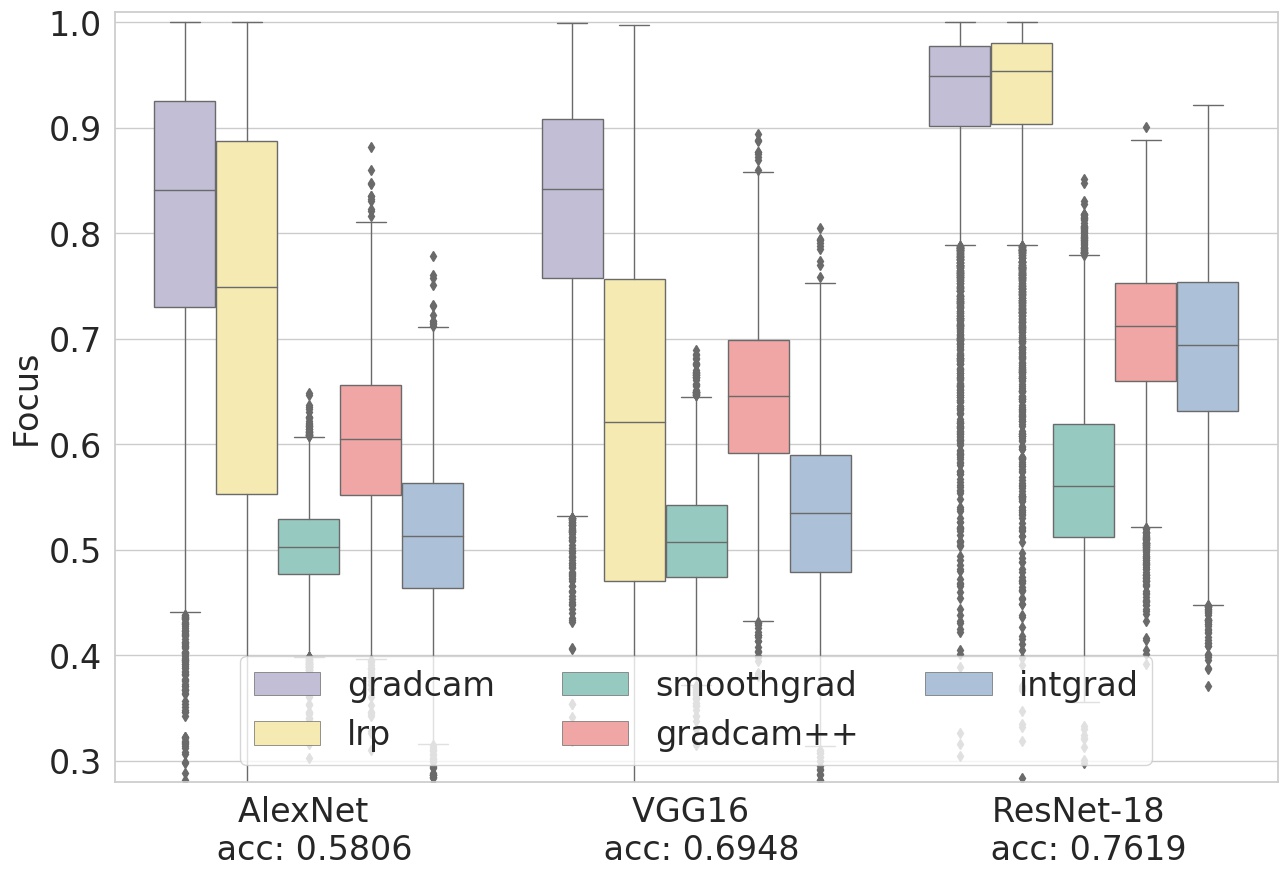} &
\includegraphics[width=0.49\textwidth]{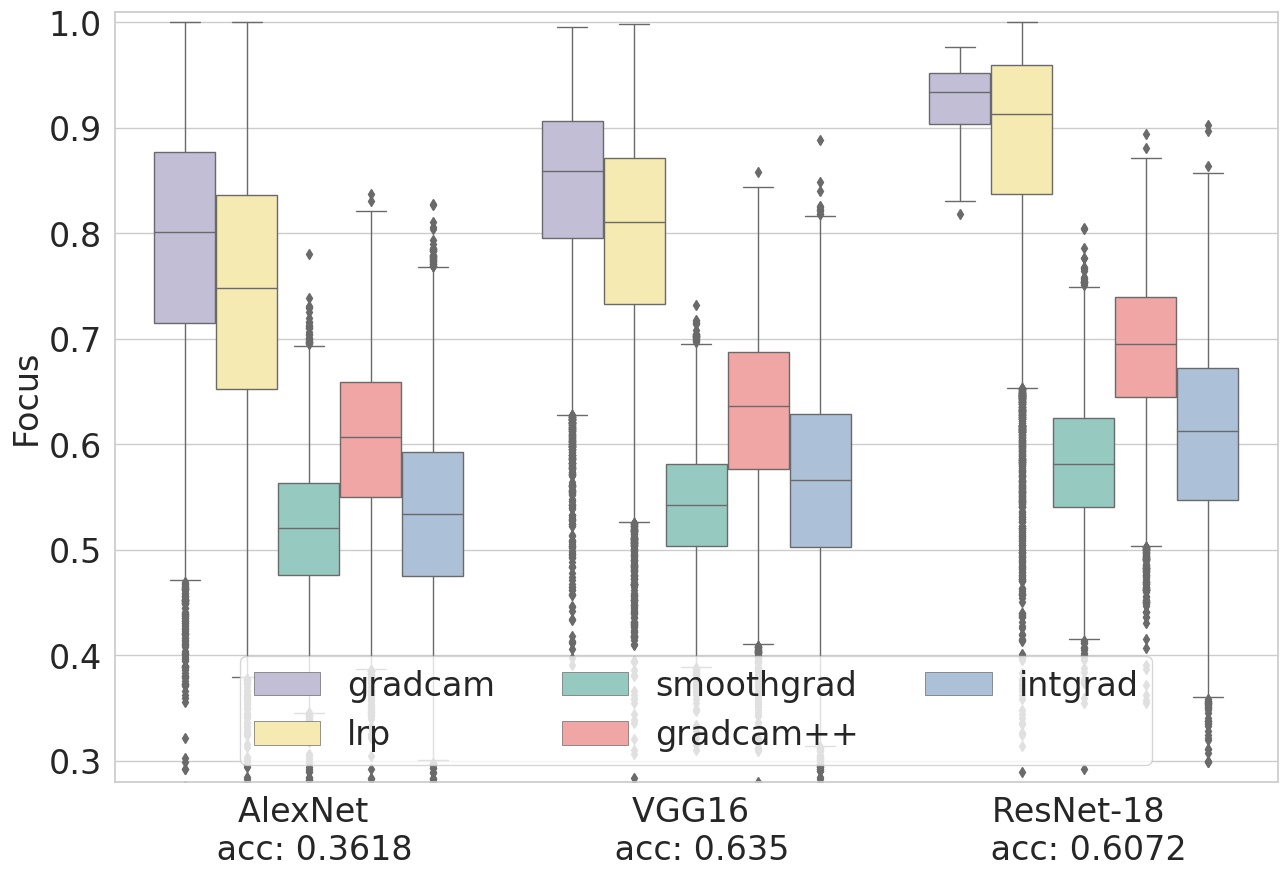} \\
(c)  & (d)  \\
\\
\end{tabular}
\caption{\focus distribution boxplot for different XAI methods applied to models trained for different datasets. The accuracy (\textit{acc}) shown under each model corresponds to the mean per class accuracy on the validation set of the corresponding dataset. These datasets are: (a) Dogs vs. Cats dataset, (b) MAMe dataset, (c) MIT67 dataset and (d) ImageNet dataset. LIME is only present in (a) and (b).}

\label{fig:all_datasets}
\end{figure*}

Let us now put \focus into practice. We evaluate GradCAM, LRP, SmoothGrad, LIME, GradCAM++ and IG, using three architectures (AlexNet, VGG16 and ResNet-18) and four target datasets (Dogs vs. Cats, MAMe, MIT67 and ImageNet). For the Dogs vs. Cats dataset, the MAMe dataset and the MIT67 dataset we use 100 mosaics per target class, a total of 200, 2,900 and 6,700 mosaics respectively. In the ImageNet experiments a total of 10,000 mosaics are used (10 per target class). Since the LIME method is computationally expensive, the experiments with this method have been restricted to the Dogs vs. Cats (200 mosaics) and MAMe datasets (2,900 mosaics). For each experiment, Table \ref{tab:results} depicts the mean and the standard deviation of the \focus distribution. For further insides, Figure \ref{fig:all_datasets} shows these distributions as box plots.

Overall, \focus seems to be correlated with model accuracy. As models get better, the mean \focus goes up and the standard deviation goes down. However, there are exceptions to this rule, as the ResNet-18 outperforms the \focus of others consistently. This indicates that certain architectures produce more precise explanations than others.

GradCAM results are the best in average. Reaching a mean \focus above 81\% in all experiments but one, it is best in 2/3 of the experiments conducted. This XAI method is particularly robust to noisy models, performing competitively even with 36\% accuracy models (AlexNet on ImageNet). GradCAM++ scores significantly lower in every experiment we conducted, being the 3rd or 4th in the overall ranking. Still, its explanations are well above random behavior.

LRP gets the second best \focus in 8 of 12 experiments, and wins in 3 of the remaining 4. As LIME, performs very well on the high accuracy models of Dogs vs. Cats, outperforming GradCAM. But on the other models it is able to beat the mean of GradCAM only once, while variance grows significantly. The worst results of LRP are produced in the MIT67 experiment, for the AlexNet and VGG16 models. Notice these models where pre-trained on the Places365-Standard dataset \cite{zhou2017places}, which is noticeably narrower than ImageNet (434 vs 1,000 classes). According to these results, LRP is a very good methodology for XAI, maybe the best, when applied to very accurate models.

LIME performs remarkably well for the Dogs vs. Cats models, 
the ones with the highest accuracy (pre-trained with ImageNet), and the only two-class classification task. For lower accuracy models (AlexNet in this task, and all in MAMe task), LIME becomes less reliable. Its mean \focus drops, and its standard deviation becomes the largest of all XAI methods. The lack of hyperparameter tuning (which is impractical) may have penalized the results for MAMe. 

SmoothGrad generally obtains a \focus around 50$\%$, showing close to random precision in all experiments. Since this method uses the gradient of the output \wrt to the input pixels, misleading attribution scores could be caused by discontinuous gradients or by saturation of gradients, as previously suggested \cite{shrikumar2017learning}. The IG method tries to overcome these drawbacks and, while its mean score is always better than the SmoothGrad, it remains quasi random in general. The cause behind these noisy explanations may be the domination of gradients in saturated areas, as shown by Miglani \etal\cite{miglani2020investigating}.

\setcounter{topnumber}{2}

\begin{table*}[!htp]
    \caption{Mean and standard deviation (in parenthesis) of the Focus distribution obtained by different XAI methods (columns) on architectures trained for different datasets (rows). The accuracy shown besides each model (\textit{acc}) corresponds to the mean per class accuracy on the validation set. Best mean Focus per row in bold.}\label{tab:results}
    \centering
    \resizebox{\linewidth}{!}{%
    \begin{tabular}{cccccccc}
 &
  \multicolumn{1}{l}{} &
  GradCAM &
  LRP &
  SmoothGrad &
  GradCAM++ &
  IntGrad &
  LIME \\ \hline
\multicolumn{1}{c}{\multirow{3}{*}{Dogs vs. Cats}} &
  AlexNet - acc: 0.9644 &
  0.9101 (± 0.0903) &
  \textbf{0.9230 (± 0.1018)} &
  0.5092 (± 0.0840) &
  0.7041 (± 0.0872) &
  0.5113 (± 0.0858) &
  0.8883 (± 0.1797) \\ 
\multicolumn{1}{c}{} &
  VGG16 - acc: 0.9893 &
  0.9446 (± 0.0577) &
  0.9526 (± 0.0877) &
  0.5035 (± 0.0854) &
  0.7574 (± 0.0777) &
  0.5108 (± 0.0849) &
  \textbf{0.9724 (± 0.1024)} \\ 
\multicolumn{1}{c}{} &
  ResNet-18  acc: 0.9878 &
  0.9725 (± 0.0320) &
  \textbf{0.9741 (± 0.1018)} &
  0.4970 (± 0.0677) &
  0.7484 (± 0.0456) &
  0.5037 (± 0.0976) &
  0.9735 (± 0.0809) \\ \hline
\multicolumn{1}{c}{\multirow{3}{*}{MAMe}} &
  AlexNet - acc: 0.7676 &
  \textbf{0.8292 (± 0.1346)} &
  0.7237 (± 0.2359) &
  0.4962 (± 0.0515) &
  0.6117 (± 0.0879) &
  0.5138 (± 0.0825) &
  0.6695 (± 0.2819) \\ 
\multicolumn{1}{c}{} &
  VGG16 - acc: 0.8069 &
  \textbf{0.8556 (± 0.1123)} &
  0.7827 (± 0.2015) &
  0.4957 (± 0.0626) &
  0.6401 (± 0.0932) &
  0.5354 (± 0.1050) &
  0.7951 (± 0.2459) \\ 
\multicolumn{1}{c}{} &
  ResNet-18 - acc: 0.8220 &
  \textbf{0.8941 (± 0.0938)} &
  0.8864 (± 0.1268) &
  0.5257 (± 0.0521) &
  0.6874 (± 0.0665) &
  0.6076 (± 0.1213) &
  0.7937 (± 0.2533) \\ \hline
\multicolumn{1}{c}{\multirow{3}{*}{MIT67}} &
  AlexNet - acc: 0.5806 &
  \textbf{0.8133 (± 0.1401)} &
  0.6864 (± 0.2545) &
  0.5017 (± 0.0415) &
  0.6037 (± 0.0773) &
  0.5121 (± 0.0736) &
  --- \\ 
\multicolumn{1}{c}{} &
  VGG16 - acc: 0.6948 &
  \textbf{0.8230 (± 0.1088)} &
  0.6033 (± 0.1978) &
  0.5079 (± 0.0522) &
  0.6441 (± 0.0776) &
  0.5340 (± 0.0809) &
  --- \\ 
\multicolumn{1}{c}{} &
  ResNet-18 - acc: 0.7619 &
  \textbf{0.9248 (± 0.0818)} &
  0.9162 (± 0.1265) &
  0.5682 (± 0.0807) &
  0.7027 (± 0.0702) &
  0.6892 (± 0.0865) &
  --- \\ \hline
\multicolumn{1}{c}{\multirow{3}{*}{ImageNet}} &
  AlexNet - acc: 0.3618 &
  \textbf{0.7866 (± 0.1179)} &
  0.7345 (± 0.1442) &
  0.5194 (± 0.0644) &
  0.6018 (± 0.0797) &
  0.5342 (±0.0867) &
  --- \\ 
\multicolumn{1}{c}{} &
  VGG16 - acc: 0.6350 &
  \textbf{0.8426 (± 0.0881)} &
  0.7914 (± 0.1140) &
  0.5425 (± 0.0566) &
  0.6279  (± 0.0814) &
  0.5637 (± 0.0924) &
  --- \\ 
\multicolumn{1}{c}{} &
  ResNet-18 - acc: 0.6072 &
  0.8792 (± 0.0849) &
  \textbf{0.8814 (± 0.1068)} &
  0.5827 (± 0.0608) &
  0.6885 (± 0.0711) &
  0.6081 (± 0.0897) &
  --- \\
  \\
\end{tabular}

    }
\end{table*}

\section{Bias detection}
\label{sec:bias_detection}

Explainability has been used to validate biases in models before. For example, the GradCAM authors use their method to visually validate the existence of gender bias in a model \cite{Selvaraju_2019}. However, this approach typically relies on a human identifying the bias beforehand. With \focus we can go beyond, automating the bias identification process as well, while providing visual validation to the user. This is possible because mosaics induce in-distribution noise, where \focus errors directly correspond to visual biases of the model.

In this section we illustrate how mosaics and \focus together can be used to identify sources of bias in a model. The proposed procedure is as follows. First, for a better detection of biases between pairs of classes, we use mosaics with two classes. Therefore, in the mosaics used for this section, samples different from the target class actually belong to the same class: $c(img_3) = c(img_4) \neq c(m)$. We concentrate on the most relevant biases by finding the pairs of classes obtaining the lowest mean \textit{Focus} in their joint mosaics. For each of these pairs we extract the mosaics with highest and lowest \textit{Focus}, and present them to a human evaluator who must review the explanations produced. The role of the evaluator is to interpret the rationale behind the explanations (both correct and incorrect) and its degree of generalization for the task. Based on that assessment, corrective measures can be implemented, as later discussed.

\begin{figure*}[t]
\centering
\setlength\tabcolsep{2pt}
\begin{tabular}{cc}
\includegraphics[width=0.48\textwidth]{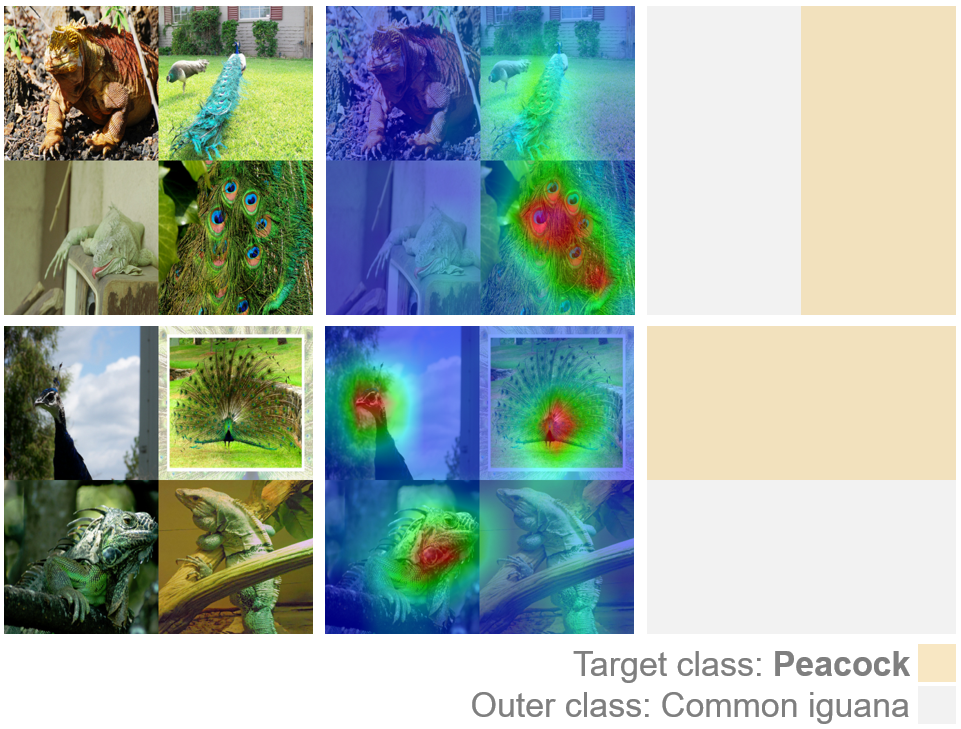} &
\includegraphics[width=0.48\textwidth]{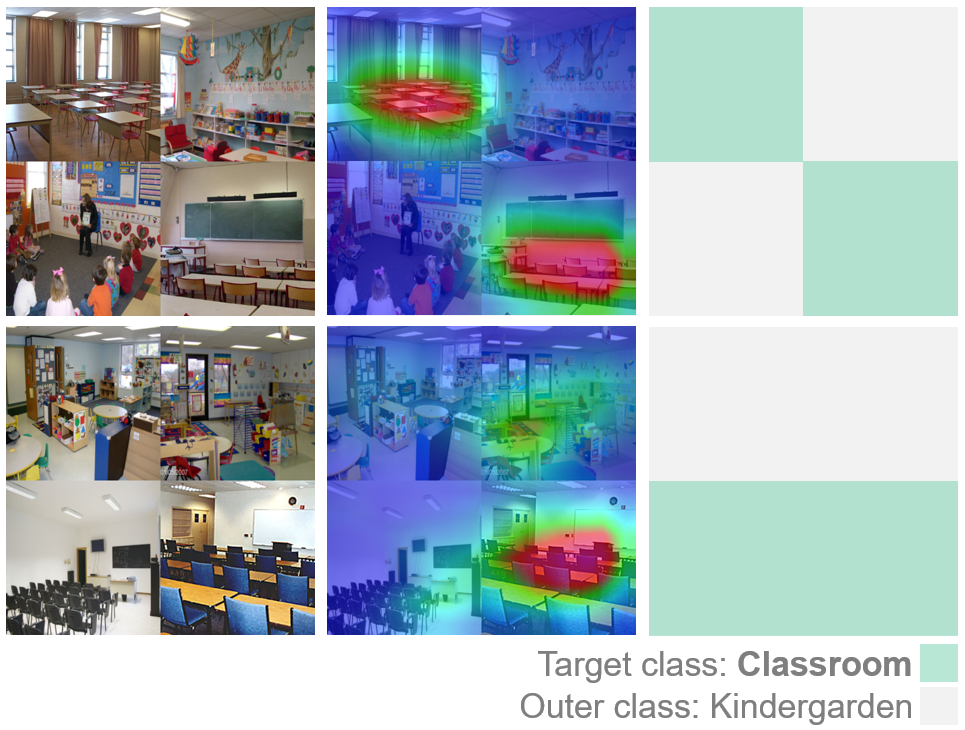} \\
(a)  & (b) \\
\\
\end{tabular}
\caption{GradCAM explanations obtained on the ResNet-18 trained with (a) ImageNet and (b) MIT67. Two examples of mosaics are shown in the first column. The second column shows the corresponding GradCAM explanations for the target class.  The third column specifies the positions of the classes within the mosaic. (a) The target class is the \textit{Peacock} class and the outer class is the \textit{Common iguana} class. The example above obtains a high \focus score (0.8176) and  the one below a lower one (0.4940). (b) The target class is the \textit{Classroom} class and the outer class the \textit{Kindergarden} class. The \focus scores are 0.8474 and 0.5960 respectively.}
\label{fig:ilsvrc2012_mit67_bias}
\end{figure*}

To conduct this experiment we use the GradCAM method and the ResNet-18 architecture, a configuration which obtains a particularly robust \textit{Focus}. A few selected examples are shown in Figure \ref{fig:ilsvrc2012_mit67_bias}, divided in two rows: the top one corresponds to a high \focus and the bottom one to a low \textit{Focus}. For the example from the ImageNet dataset (see Figure \ref{fig:ilsvrc2012_mit67_bias} (a)), the model is able to correctly attribute relevance to the \textit{Peacock} images on the upper mosaic, while, for the bottom mosaic, some of the relevance incorrectly fall on the head of the \textit{Common iguana}. The fact that most of the incorrect relevance in the \textit{Common iguana} falls in the subtympanic shield (\ie the characteristic circle in its jowl) seems to be related with its visual similarity with the ocellus of the \textit{Peacock} (\ie the circular spot in the feathers). Notice the iguana's subtympanic shield is hardly visible in the top mosaic. For the example from the MIT67 dataset (see Figure \ref{fig:ilsvrc2012_mit67_bias} (b)), the model correctly attributes the relevance to the two target class images on the top mosaic, both belonging to the \textit{Classroom} class. For the lower mosaic, the model struggles to find the evidence in the \textit{Classroom} image when no tables are present. These patterns are consistent found in several mosaics for the classes studied.

After reviewing several cases as the ones described above, one can identify at least two types of biases responsible for decreasing the \focus score. These are: 
\begin{enumerate}
    \item Shared bias: A visual evidence of the target class is found in an outer class image (\eg the ocellus shape found in the \textit{Common iguana} class).
    \item Missing bias: A visual evidence of the target class is not found in an image of the same target class (\eg the tables in the \textit{Classroom} class).
\end{enumerate}

After the identification of these biases, and an assessment of their impact, one could try to mitigate their relevance for the model. For casuistry (1), shared bias, more images of the target class without the characteristic pattern found in the outer class could be added to the training set (\eg \textit{Peacocks} images where the ocellus is not visible). Similarly, more images of the outer class where the characteristic pattern is present (\eg \textit{Common iguana} images where the subtympanic shield is visible) could be added. In either case, the dependency of the target class \wrt the shared bias would be reduced, increasing the robustness of the model. For casuistry (2), missing bias, more samples without the identified visual patterns of the target class could be added to the training set (\eg \textit{Classrooms} samples without tables). Again, this would reduce the dependency of the target class \wrt the missing bias.

\section{Conclusion}
\label{sec:conclusion}

With the aim of evaluating XAI methods in a quantitative manner, we introduce a novel metric---the \textit{Focus}---to assess the faithfulness of a XAI method to the underlying model. We show the methodology to be consistent and resilient to misleading explanations.

When applied to SmoothGrad or IG, \focus finds these methodologies as quasi-random in its explanations \wrt the model. On the contrary, LRP and GradCAM are both found to be consistently reliable methods. GradCAM performs well on all experiments conducted, even when the underlying model is not particularly well fit to the task. LRP performs very well for high performing models, but it becomes more unreliable on less accurate models. This also seems to be the case of LIME, which suffers from an even larger variance. Furthermore, LIME computational complexity and need for hyperparameter tuning limits its practical application. GradCAM++ performs better than random, but not as well as GradCAM and LRP. Remarkably, the \focus results are rather consistent across tasks and architectures, providing strong empirical evidence of their performance.

The consistency of \focus is likely related with the type of noise it induces. By altering the context and not the content of samples, \focus adds and exploits in-distribution noise. Unlike out-distribution noise, this is less prone to arbitrary model behavior. Through in-distribution noise mosaics and the \focus score visually characterize bias in the model, and can be directly used as an automated bias identification and exemplification tool. This opens the door to use mosaics and \focus to improve models, datasets and explanations.

\focus is related with the precision metric (\ie $\frac{TP}{TP+FP}$). While \focus is not precision (it lacks the ground truth needed to specify TP from FP), it approximates it by implicit labeling of mosaic quadrants. That is, that all positive lays somewhere within the target quadrants, and all negative somewhere within the other two quadrants. A similar assumption could be made to define an analogous recall metric (\ie $\frac{TP}{TP+FN}$) using, for example, the negative relevance provided by some \textit{feature attribution} methods. This remains as future work.

\section*{Acknowledgements}
This work is supported by the European Union – Horizon 2020 Program under the scheme “INFRAIA-01-2018-2019 – Integrating Activities for Advanced Communities”, Grant Agreement n.871042, “SoBigData++: European Integrated Infrastructure for Social Mining and Big Data Analytics” (http://www.sobigdata.eu) and by the Departament de Recerca i Universitats of the Generalitat de Catalunya under the Industrial Doctorate Grant DI 2018-100. We thank Marta González Mallo for the insightful comments and suggestions.

\bibliographystyle{unsrtnat}
\bibliography{focus}  

\begin{thebibliography}{25}
\providecommand{\natexlab}[1]{#1}
\providecommand{\url}[1]{\texttt{#1}}
\expandafter\ifx\csname urlstyle\endcsname\relax
  \providecommand{\doi}[1]{doi: #1}\else
  \providecommand{\doi}{doi: \begingroup \urlstyle{rm}\Url}\fi

\bibitem[Gilpin et~al.(2018)Gilpin, Bau, Yuan, Bajwa, Specter, and
  Kagal]{gilpin2018explaining}
Leilani~H Gilpin, David Bau, Ben~Z Yuan, Ayesha Bajwa, Michael Specter, and
  Lalana Kagal.
\newblock Explaining explanations: An overview of interpretability of machine
  learning.
\newblock In \emph{2018 IEEE 5th International Conference on data science and
  advanced analytics (DSAA)}, pages 80--89. IEEE, 2018.

\bibitem[Mohseni et~al.(2018)Mohseni, Zarei, and
  Ragan]{mohseni2018multidisciplinary}
Sina Mohseni, Niloofar Zarei, and Eric~D Ragan.
\newblock A multidisciplinary survey and framework for design and evaluation of
  explainable ai systems.
\newblock \emph{arXiv preprint arXiv:1811.11839}, 2018.

\bibitem[Selvaraju et~al.(2019)Selvaraju, Cogswell, Das, Vedantam, Parikh, and
  Batra]{Selvaraju_2019}
Ramprasaath~R. Selvaraju, Michael Cogswell, Abhishek Das, Ramakrishna Vedantam,
  Devi Parikh, and Dhruv Batra.
\newblock Grad-cam: Visual explanations from deep networks via gradient-based
  localization.
\newblock \emph{International Journal of Computer Vision}, 128\penalty0
  (2):\penalty0 336–359, Oct 2019.
\newblock ISSN 1573-1405.
\newblock \doi{10.1007/s11263-019-01228-7}.
\newblock URL \url{http://dx.doi.org/10.1007/s11263-019-01228-7}.

\bibitem[Bach et~al.(2015)Bach, Binder, Montavon, Klauschen, M{\"u}ller, and
  Samek]{bach2015pixel}
Sebastian Bach, Alexander Binder, Gr{\'e}goire Montavon, Frederick Klauschen,
  Klaus-Robert M{\"u}ller, and Wojciech Samek.
\newblock On pixel-wise explanations for non-linear classifier decisions by
  layer-wise relevance propagation.
\newblock \emph{PloS one}, 10\penalty0 (7):\penalty0 e0130140, 2015.

\bibitem[Chattopadhay et~al.(2018)Chattopadhay, Sarkar, Howlader, and
  Balasubramanian]{chattopadhay2018grad}
Aditya Chattopadhay, Anirban Sarkar, Prantik Howlader, and Vineeth~N
  Balasubramanian.
\newblock Grad-cam++: Generalized gradient-based visual explanations for deep
  convolutional networks.
\newblock In \emph{2018 IEEE winter conference on applications of computer
  vision (WACV)}, pages 839--847. IEEE, 2018.

\bibitem[Ribeiro et~al.(2016)Ribeiro, Singh, and Guestrin]{ribeiro2016should}
Marco~Tulio Ribeiro, Sameer Singh, and Carlos Guestrin.
\newblock " why should i trust you?" explaining the predictions of any
  classifier.
\newblock In \emph{Proceedings of the 22nd ACM SIGKDD international conference
  on knowledge discovery and data mining}, pages 1135--1144, 2016.

\bibitem[Sundararajan et~al.(2017)Sundararajan, Taly, and
  Yan]{sundararajan2017axiomatic}
Mukund Sundararajan, Ankur Taly, and Qiqi Yan.
\newblock Axiomatic attribution for deep networks.
\newblock In \emph{International Conference on Machine Learning}, pages
  3319--3328. PMLR, 2017.

\bibitem[Adebayo et~al.(2018)Adebayo, Gilmer, Muelly, Goodfellow, Hardt, and
  Kim]{adebayo2020sanity}
Julius Adebayo, Justin Gilmer, Michael Muelly, Ian Goodfellow, Moritz Hardt,
  and Been Kim.
\newblock Sanity checks for saliency maps.
\newblock \emph{arXiv preprint arXiv:1810.03292}, 2018.

\bibitem[Zhang et~al.(2018)Zhang, Bargal, Lin, Brandt, Shen, and
  Sclaroff]{zhang2018top}
Jianming Zhang, Sarah~Adel Bargal, Zhe Lin, Jonathan Brandt, Xiaohui Shen, and
  Stan Sclaroff.
\newblock Top-down neural attention by excitation backprop.
\newblock \emph{International Journal of Computer Vision}, 126\penalty0
  (10):\penalty0 1084--1102, 2018.

\bibitem[Samek and M{\"u}ller(2019)]{Samek_2019}
Wojciech Samek and Klaus-Robert M{\"u}ller.
\newblock Towards explainable artificial intelligence.
\newblock In \emph{Explainable AI: interpreting, explaining and visualizing
  deep learning}, pages 5--22. Springer, 2019.

\bibitem[Nam et~al.(2019)Nam, Gur, Choi, Wolf, and Lee]{nam2019relative}
Woo-Jeoung Nam, Shir Gur, Jaesik Choi, Lior Wolf, and Seong-Whan Lee.
\newblock Relative attributing propagation: Interpreting the comparative
  contributions of individual units in deep neural networks, 2019.

\bibitem[Montavon et~al.(2017)Montavon, Lapuschkin, Binder, Samek, and
  M{\"u}ller]{montavon2017explaining}
Gr{\'e}goire Montavon, Sebastian Lapuschkin, Alexander Binder, Wojciech Samek,
  and Klaus-Robert M{\"u}ller.
\newblock Explaining nonlinear classification decisions with deep taylor
  decomposition.
\newblock \emph{Pattern Recognition}, 65:\penalty0 211--222, 2017.

\bibitem[Smilkov et~al.(2017)Smilkov, Thorat, Kim, Vi{\'e}gas, and
  Wattenberg]{smilkov2017smoothgrad}
Daniel Smilkov, Nikhil Thorat, Been Kim, Fernanda Vi{\'e}gas, and Martin
  Wattenberg.
\newblock Smoothgrad: removing noise by adding noise.
\newblock \emph{arXiv preprint arXiv:1706.03825}, 2017.

\bibitem[Kokhlikyan et~al.(2020)Kokhlikyan, Miglani, Martin, Wang, Alsallakh,
  Reynolds, Melnikov, Kliushkina, Araya, Yan, and
  Reblitz-Richardson]{kokhlikyan2020captum}
Narine Kokhlikyan, Vivek Miglani, Miguel Martin, Edward Wang, Bilal Alsallakh,
  Jonathan Reynolds, Alexander Melnikov, Natalia Kliushkina, Carlos Araya, Siqi
  Yan, and Orion Reblitz-Richardson.
\newblock Captum: A unified and generic model interpretability library for
  pytorch, 2020.

\bibitem[Krizhevsky et~al.(2012)Krizhevsky, Sutskever, and
  Hinton]{krizhevsky2012imagenet}
Alex Krizhevsky, Ilya Sutskever, and Geoffrey~E Hinton.
\newblock Imagenet classification with deep convolutional neural networks.
\newblock \emph{Advances in neural information processing systems},
  25:\penalty0 1097--1105, 2012.

\bibitem[Simonyan and Zisserman(2014)]{simonyan2014very}
Karen Simonyan and Andrew Zisserman.
\newblock Very deep convolutional networks for large-scale image recognition.
\newblock \emph{arXiv preprint arXiv:1409.1556}, 2014.

\bibitem[He et~al.(2016)He, Zhang, Ren, and Sun]{he2016deep}
Kaiming He, Xiangyu Zhang, Shaoqing Ren, and Jian Sun.
\newblock Deep residual learning for image recognition.
\newblock In \emph{Proceedings of the IEEE conference on computer vision and
  pattern recognition}, pages 770--778, 2016.

\bibitem[Parés et~al.(2021)Parés, Arias-Duart, Garcia-Gasulla,
  Campo-Francés, Viladrich, Ayguadé, and Labarta]{pares2021mame}
Ferran Parés, Anna Arias-Duart, Dario Garcia-Gasulla, Gema Campo-Francés,
  Nina Viladrich, Eduard Ayguadé, and Jesús Labarta.
\newblock The mame dataset: On the relevance of high resolution and variable
  shape image properties, 2021.

\bibitem[Quattoni and Torralba(2009)]{quattoni2009recognizing}
Ariadna Quattoni and Antonio Torralba.
\newblock Recognizing indoor scenes.
\newblock In \emph{2009 IEEE Conference on Computer Vision and Pattern
  Recognition}, pages 413--420. IEEE, 2009.

\bibitem[Russakovsky et~al.(2015)Russakovsky, Deng, Su, Krause, Satheesh, Ma,
  Huang, Karpathy, Khosla, Bernstein, et~al.]{russakovsky2015imagenet}
Olga Russakovsky, Jia Deng, Hao Su, Jonathan Krause, Sanjeev Satheesh, Sean Ma,
  Zhiheng Huang, Andrej Karpathy, Aditya Khosla, Michael Bernstein, et~al.
\newblock Imagenet large scale visual recognition challenge.
\newblock \emph{International journal of computer vision}, 115\penalty0
  (3):\penalty0 211--252, 2015.

\bibitem[Reddi et~al.(2019)Reddi, Kale, and Kumar]{reddi2019convergence}
Sashank~J Reddi, Satyen Kale, and Sanjiv Kumar.
\newblock On the convergence of adam and beyond.
\newblock \emph{arXiv preprint arXiv:1904.09237}, 2019.

\bibitem[Zhou et~al.(2017)Zhou, Lapedriza, Khosla, Oliva, and
  Torralba]{zhou2017places}
Bolei Zhou, Agata Lapedriza, Aditya Khosla, Aude Oliva, and Antonio Torralba.
\newblock Places: A 10 million image database for scene recognition.
\newblock \emph{IEEE Transactions on Pattern Analysis and Machine
  Intelligence}, 2017.

\bibitem[Szegedy et~al.(2016)Szegedy, Vanhoucke, Ioffe, Shlens, and
  Wojna]{szegedy2016rethinking}
Christian Szegedy, Vincent Vanhoucke, Sergey Ioffe, Jon Shlens, and Zbigniew
  Wojna.
\newblock Rethinking the inception architecture for computer vision.
\newblock In \emph{Proceedings of the IEEE conference on computer vision and
  pattern recognition}, pages 2818--2826, 2016.

\bibitem[Shrikumar et~al.(2017)Shrikumar, Greenside, and
  Kundaje]{shrikumar2017learning}
Avanti Shrikumar, Peyton Greenside, and Anshul Kundaje.
\newblock Learning important features through propagating activation
  differences.
\newblock In \emph{International Conference on Machine Learning}, pages
  3145--3153. PMLR, 2017.

\bibitem[Miglani et~al.(2020)Miglani, Kokhlikyan, Alsallakh, Martin, and
  Reblitz-Richardson]{miglani2020investigating}
Vivek Miglani, Narine Kokhlikyan, Bilal Alsallakh, Miguel Martin, and Orion
  Reblitz-Richardson.
\newblock Investigating saturation effects in integrated gradients.
\newblock \emph{arXiv preprint arXiv:2010.12697}, 2020.

\end{thebibliography}

\end{document}